\documentclass{article}

\PassOptionsToPackage{numbers, compress}{natbib}
\usepackage[final]{nips_2016}

\usepackage[utf8]{inputenc} % allow utf-8 input
\usepackage[T1]{fontenc}    % use 8-bit T1 fonts
\usepackage{hyperref}       % hyperlinks
\usepackage{url}            % simple URL typesetting
\usepackage{booktabs}       % professional-quality tables
\usepackage{amsfonts}       % blackboard math symbols
\usepackage{nicefrac}       % compact symbols for 1/2, etc.
\usepackage{microtype}      % microtypography
\usepackage{amsmath}
\usepackage{graphicx}
\usepackage[utf8]{inputenc}

\DeclareMathOperator*{\argmin}{argmin}

\title{Deep Directed Generative Models\\with Energy-Based Probability Estimation}

\author{
Taesup Kim, Yoshua Bengio\thanks{CIFAR Senior Member } \\
Department of Computer Science and Operations Research \\
Université de Montréal\\
Montréal, QC, Canada\\
\texttt{\{taesup.kim,yoshua.bengio\}@umontreal.ca} \\
}

\begin{document}

\maketitle

\begin{abstract}
Training energy-based probabilistic models is confronted with apparently intractable sums, whose Monte Carlo estimation requires sampling from the estimated probability distribution in the inner loop of training. 
This can be approximately achieved by Markov chain Monte Carlo methods, but may still face a formidable obstacle that is the difficulty of mixing between modes with sharp concentrations of probability. 
Whereas an MCMC process is usually derived from a given energy function based on mathematical considerations and requires an arbitrarily long time to obtain good and varied samples, we propose to train a deep directed generative model (not a Markov chain) so that its sampling distribution approximately matches the energy function that is being trained.
Inspired by generative adversarial networks, the proposed framework involves training of two models that represent dual views of the estimated probability distribution: the energy function (mapping an input configuration to a scalar energy value) and the generator (mapping a noise vector to a generated configuration), both represented by deep neural networks.
\end{abstract}

\section{Introduction}\label{SEC:INTRO}
Energy-based models capture dependencies over random variables of interest by defining an energy function, and the probability distribution can be further obtained by normalizing the exponentiated and negated energy function. 
The energy function associates each configuration of random variables with a scalar energy value, and lower energy values are assigned to more likely or plausible configurations. 
The energy function is typically used in order to parametrize undirected graphical models, such as Boltzmann machines~\cite{BM}, in the form of a Boltzmann distribution with an appropriate normalization factor. 
In general, the normalization factor introduces some difficulties during the maximum likelihood training of energy-based models, because it is a sum over all configurations of random variables and the corresponding gradient is an average of the energy gradient for configurations sampled from the model itself.
Not only is this sum intractable, but exact Monte Carlo sampling from it is also intractable. 
In order to estimate the gradients of the normalization factor, Markov chain Monte Carlo (MCMC) methods are typically used to obtain approximate samples from the model distribution.
However, it is usual that MCMC methods make small moves as probable that are very unlikely to jump between separate modes
As the model distribution becomes sharper with multiple modes separated by very low probability regions during the training, the difficulty of sampling from MCMC methods arises in this context~\cite{MIXING}.
In order to sidestep this problem, we propose to {\em train} a deep directed generative model that produces a sample by deterministically transforming an independent and identically distributed (i.i.d.) random sample, such as a uniform variate. 
This avoids the need of a sequential process with an arbitrarily long computation time to generate samples for training energy-based probabilistic models. 
In the proposed framework, the learned knowledge is now represented through two complementary models and dual views: the energy function and the generator. 
The energy function is trained in a way estimating the maximum likelihood gradient that the approximate samples from the model (needed to estimate the gradient of the normalization factor) are obtained from the generator model rather than from a Markov chain. 
The generator is trained in a similar way as generative adversarial networks (GAN)~\cite{GAN}, i.e., the energy function can be considered as a discriminator: low energy corresponds to "real" data (because the energy function is trained to assign low energy on training examples) and high energy to "fake" or generated data (when the generator is putting probability mass in wrong places).
The energy function thus provides gradients that encourage the generator to produce lower energy samples. 
Since the generator of GAN suffers from a missing mode problem, we introduce a regularization term that indirectly maximizes the entropy in the training objective of the generator, which are empirically shown to be essential to obtain more plausible samples.

\section{Energy-Based Probabilistic Models}
\subsection{Products of Experts}
Energy-based models associate a scalar energy value to each configuration of random variables, which we consider it as input data $\mathbf{x}$, with an energy function $E_{\Theta}$ and a trainable parameter set $\Theta$~\cite{EBM}.
The energy function is typically trained to assign lower energy values to more plausible or desirable configurations, i.e., where the training examples are. 
Moreover, it can be extended into a probabilistic model by using a Boltzmann distribution (also called Gibbs distribution):
\begin{equation}\label{EQ:ENERGY}
P_{\Theta}(\mathbf{x})=\frac{e^{-E_{\Theta}(\mathbf{x})}}{Z_{\Theta}} \qquad Z_{\Theta}=\sum_{\mathbf{x}}{e^{-E_{\Theta}(\mathbf{x})}}
\end{equation}
where $Z_{\theta}$ is a normalization factor and also called as a partition function.
As we typically define the energy function as a sum of multiple terms, we can associate each term with an "expert", and the corresponding probability distribution is expressed as a product of experts (PoE)~\cite{POE}:
\begin{equation}\label{EQ:POE}
E_{\Theta}(\mathbf{x})=\sum_{i}{\tilde{E}_{\theta_{i}}(\mathbf{x})}\qquad
P_{\Theta}(\mathbf{x})=\frac{e^{-\sum_{i}{\tilde{E}_{\theta_{i}}(\mathbf{x})}}}{Z_{\Theta}}=\frac{1}{Z_{\Theta}}\prod_{i}e^{-\tilde{E}_{\theta_{i}}(\mathbf{x})}=\frac{1}{Z_{\Theta}}\prod_{i}\tilde{P}_{\theta_{i}}(\mathbf{x})
\end{equation}
Each expert $\tilde{E}_{\theta_{i}}$ is defined by an unnormalized distribution as $\tilde{E}_{\theta_{i}}(\mathbf{x})=-\log{\tilde{P}_{\theta_{i}}(\mathbf{x})}$ with a parameter set $\theta_{i}$, and also can be interpreted as a pattern detector of implausible configurations.
For example, in the product of Student-t (PoT) model~\cite{POT}, the energy function is defined by using a set of unnormalized distributions in the form of Student-t distribution, which has heavier tails than a normal distribution:
\begin{equation}\label{EQ:POT}
\begin{gathered}
\tilde{P}_{\theta_{i}}(\mathbf{x})=\frac{1}{\big(1+\frac{1}{2}(W_{i}^{T}\mathbf{x})^{2}\big)^{\alpha_{i}}} \quad \alpha_{i}>0\\
E_{\Theta}(x)=\sum_{i}{\alpha_{i}\log{\Big(1+\big(W_{i}^{T}x\big)^{2}\Big)}}
\end{gathered}
\end{equation}
It is also possible to use weak classifiers based on logistic regression as unnormalized distributions. Each weak classifier corresponds to a single feature detector, and the responses of all weak classifier are aggregated to assign an energy value: 
\begin{equation}\label{EQ:LOGIT}
\begin{gathered}
\tilde{P}_{\theta_{i}}(\mathbf{x})=\sigma{\big(W_{i}^{T}\mathbf{x}+b_{i}\big)}\\
E_{\Theta}(\mathbf{x})=\sum_{i}{\log{\big(1+e^{-(W_{i}^{T}\mathbf{x}+b_{i})}\big)}}
\end{gathered}
\end{equation}
Interestingly, restricted Boltzmann machines (RBM)~\cite{RBM} have a free energy over Gaussian visible units $\mathbf{x}$ with terms associated with binary hidden units $\mathbf{h}$ corresponding to experts that try to classify the input data $\mathbf{x}$ as being "real" or "fake" data, and the energy gets low when all experts agree:
\begin{equation}\label{EQ:RBM}
\begin{aligned}
E_{\Theta}(\mathbf{x})&=\frac{1}{\sigma^{2}}\mathbf{x}^{T}\mathbf{x}-b^{T}\mathbf{x}-\sum_{i}{\log{\sum_{h_{i}}{e^{h_{i}(W_{i}^{T}\mathbf{x}+b_{i})}}}}\\
&=\frac{1}{\sigma^{2}}\mathbf{x}^{T}\mathbf{x}-b^{T}\mathbf{x}-\sum_{i}{\log{\big(1+e^{W_{i}^{T}\mathbf{x}+b_{i}}}\big)}
\end{aligned}
\end{equation}

\subsection{Learning and Intractability}\label{SEC:LEARNING}
In order to estimate the data distribution $P_{\mathcal{D}}$, which is the target distribution, by using energy-based probabilistic models, the energy model distribution $P_{\Theta}$ is trained to approach $P_{\mathcal{D}}$ as much as possible.
This is done by minimizing the Kullback-Leibler (KL) divergence between the two distributions $\mathrm{D}_{KL}\big(P_{\mathcal{D}}(\mathbf{x})||P_{\Theta}(\mathbf{x})\big)$, and it corresponds to the maximum likelihood objective:
\begin{equation}\label{EQ:KL}
\Theta^{*}=\argmin_{\Theta}{\mathrm{D}_{KL}\big(P_{\mathcal{D}}(\mathbf{x})||P_{\Theta}(\mathbf{x})\big)}
=\argmin_{\Theta}{\mathrm{E}_{\mathbf{x}\sim P_{\mathcal{D}}(\mathbf{x})}\big[-\log{P_{\Theta}(\mathbf{x})}\big]}
\end{equation}
With this training criterion, we define the loss function $\mathcal{L}(\Theta, \mathcal{D}')=-\frac{1}{N}\sum_{i=1}^{N}\log{P_{\Theta}(\mathbf{x}^{(i)})}$ with the training dataset $\mathcal{D}'=\{\mathbf{x^{(i)}}\}^{N}_{i=1}$, which we assume samples are drawn from the data distribution.
Moreover, we see that it is an approximation of Eq.~\ref{EQ:KL} based on using the empirical distribution instead of the actual data distribution. 
This loss can be minimized by gradient-based optimization methods, and the parameter set $\Theta$ is updated with the following gradient:
\begin{equation}\label{EQ:POS_NEG_PHASES}
\begin{aligned}
\dfrac{\partial\mathcal{L}(\Theta, \mathcal{D}')}{\partial\Theta}&=-\dfrac{1}{N}\sum_{i=1}^{N}{\dfrac{\partial\log{P_{\Theta}(\mathbf{x}^{(i)})}}{\partial\Theta}}\\
&=\dfrac{1}{N}\sum_{i=1}^{N}{\dfrac{\partial E_{\Theta}(\mathbf{x}^{(i)})}{\partial \Theta}} - \mathrm{E}_{\mathbf{x}\sim P_{\Theta}(\mathbf{x})}\Bigg[{\dfrac{\partial E_{\Theta}(\mathbf{x})}{\partial \Theta}}\Bigg]\\
&\approx\underbrace{\mathrm{E}_{\mathbf{x}^{+}\sim P_{\mathcal{D}}(\mathbf{x})}\Bigg[{\dfrac{\partial E_{\Theta}(\mathbf{x}^{+})}{\partial \Theta}}\Bigg]}_{\text{Positive Phase}}-\underbrace{\mathrm{E}_{\mathbf{x}^{-}\sim P_{\Theta}(x)}\Bigg[{\dfrac{\partial E_{\Theta}(\mathbf{x}^{-})}{\partial \Theta}}\Bigg]}_{\text{Negative Phase}}
\end{aligned}
\end{equation}
The gradient is traditionally decomposed into two different terms that are referred to as the positive and negative phase terms, respectively. 
We can consider these two terms as two distinct forces shaping the energy function. 
The positive phase term is to decrease the energy of training examples $\mathbf{x}^{+}$ (positive samples), whereas the negative phase term is to increase the energy of samples $\mathbf{x}^{-}$ (negative samples) drawn from the energy model distribution. 
If the energy model distribution matches the data distribution, the two terms are equal that the log-likelihood gradient is at a local maximum. 
The expectation in the positive phase can be computed exactly by summing over all training examples or by using Monte Carlo estimation for stochastic gradient descent. 
However, the expectation in the negative phase requires sampling based on the model associated with the energy function, and in general exact unbiased sampling is not tractable. 
Therefore, MCMC methods are typically used with an iterative sampling procedure.
If the data distribution has a complex distribution with many sharp modes, the training eventually reaches a point, where an MCMC sampling from the model becomes problematic taking too much time to mix between modes~\cite{Bengio+chapter2007}. 
It means the ability to get good approximate model samples is an important ingredient in training energy-based probabilistic models. 
This motivated this work to {\em train a deep directed generative model whose sampling distribution approximates the energy model distribution} that it doesn't require an MCMC but a simple computation such as transforming an i.i.d. random sample (the latent variable) with a neural network.

\subsection{Training Models as a Classification Problem}\label{SEC:CLASSIFICATION}
In this section, we revisit the old idea that the update rule in Sec.~\ref{SEC:LEARNING} can be viewed as learning a classifier~\cite{Bengio_AI,MAX_CLASS}. 
Let us first assume that we are training a binary classifier to separate positive samples
from negative samples, and an additional binary variable $y$ is used as a label to indicate whether a sample $\mathbf{x}$ is a positive sample $\mathbf{x}^{+}=(\mathbf{x}, y=1) \sim P_{\mathcal{D}}(\mathbf{x})$ or a negative sample $\mathbf{x}^{-}=(\mathbf{x}, y=0) \sim P_{\Theta}(\mathbf{x})$. 
Then, we define the binary classifier as $P_{\psi}(y=1|x)=\sigma{\big(-E'_{\psi}(x)\big)}$, where $\sigma{(\cdot)}$ is a sigmoid function, and $E'_{\psi}(x)$ is an unnormalized discriminating function, such as the energy function in Eq.~\ref{EQ:ENERGY}. This can be trained by minimizing the expected negative conditional log-likelihood over the joint distribution $P(\mathbf{x},y)$ with the data distribution $P_{\mathcal{D}}(\mathbf{x})=P(\mathbf{x}|y=1)$ and energy model distribution $P_{\Theta}(\mathbf{x})=P(\mathbf{x}|y=0)$ and assuming $P(y=0)=P(y=1)=\frac{1}{2}$.
The corresponding gradient with respect to the binary classifier parameter $\psi$ is written:
\begin{equation}\label{classifier}
\begin{aligned}
&\mathrm{E}_{(\mathbf{x},y)\sim P(\mathbf{x},y)}\bigg[-\frac{\partial\log{P_{\psi}(y|\mathbf{x})}}{\partial\psi}\bigg]\\
&=-\mathrm{E}_{(\mathbf{x},y)\sim P(\mathbf{x},y)}\bigg[\frac{\partial\big(\log{{P_{\psi}(y=1|\mathbf{x})}^{y}{P_{\psi}(y=0|\mathbf{x})}^{(1-y)}\big)}}{\partial\psi}\bigg]\\
&=-\frac{1}{2}\Bigg(\mathrm{E}_{\mathbf{x}^{+}\sim P_{\mathcal{D}}(\mathbf{x})}\bigg[\frac{\partial\log{P_{\psi}(y=1|\mathbf{x}^{+})}}{\partial\psi}\bigg]+\mathrm{E}_{\mathbf{x}^{-}\sim P_{\Theta}(\mathbf{x})}\bigg[\frac{\partial\log{P_{\psi}(y=0|\mathbf{x}^{-})}}{\partial\psi}\bigg]\Bigg)\\
&=\frac{1}{2}\Bigg(\mathrm{E}_{\mathbf{x}^{+}\sim P_{\mathcal{D}}(\mathbf{x})}\bigg[P_{\psi}(y=0|\mathbf{x}^{+})\frac{\partial{E'_{\psi}(\mathbf{x}^{+})}}{\partial\psi}\bigg]-\mathrm{E}_{\mathbf{x}^{-}\sim P_{\Theta}(\mathbf{x})}\bigg[P_{\psi}(y=1|\mathbf{x}^{-})\frac{\partial{E'_{\psi}(\mathbf{x}^{-})}}{\partial\psi}\bigg]\Bigg)\\
&\approx\frac{1}{4}\Bigg(\mathrm{E}_{\mathbf{x}^{+}\sim P_{\mathcal{D}}(\mathbf{x})}\bigg[\frac{\partial{E'_{\psi}(\mathbf{x}^{+})}}{\partial\psi}\bigg]-\mathrm{E}_{\mathbf{x}^{-}\sim P_{\Theta}(\mathbf{x})}\bigg[\frac{\partial{E'_{\psi}(\mathbf{x}^{-})}}{\partial\psi}\bigg]\Bigg)
\end{aligned}
\end{equation}
where the last approximation applies with an assumption $P_{\psi}(y=0|\mathbf{x})\approx P_{\psi}(y=1|\mathbf{x})$.
It means the two categories (samples from the data distribution and the model distribution) are difficult to discriminate that should be optimally true at the end of training.
Interestingly, this approximation leads to the same update rule as in Eq.~\ref{EQ:POS_NEG_PHASES}, and this motivated us to view the energy function as defining a binary classifier that separates the training examples from samples generated by a separate generative model.
\subsection{Models with Multiple Layers}
Models with multiple layers, which are called deep models, are considered to have more ability to learn rich and complex data by extracting high-level abstractions. 
However, energy-based probabilistic models with multiple stochastic hidden layers, such as deep belief networks (DBN)~\cite{DBN} and deep Boltzmann machines (DBM)~\cite{DBM}, involve difficult inference and learning, requiring the computation or sampling from the conditional posterior over stochastic hidden units, and exact sampling from these distributions is generally intractable. 
Another option is to directly define an energy function through a deep neural network, as was proposed by ~\cite{DEM}. 
In that case, the layers of the network do not represent latent variables but rather are deterministic transformations of input data, which help to ascertain whether the input configuration is plausible (low energy) or not.  
This approach eliminates the need for inference, but still involves the high-cost or high-variance MCMC sampling to estimate the gradient of the partition function.
\begin{figure}[t]
\centering
\includegraphics[width=0.8\textwidth]{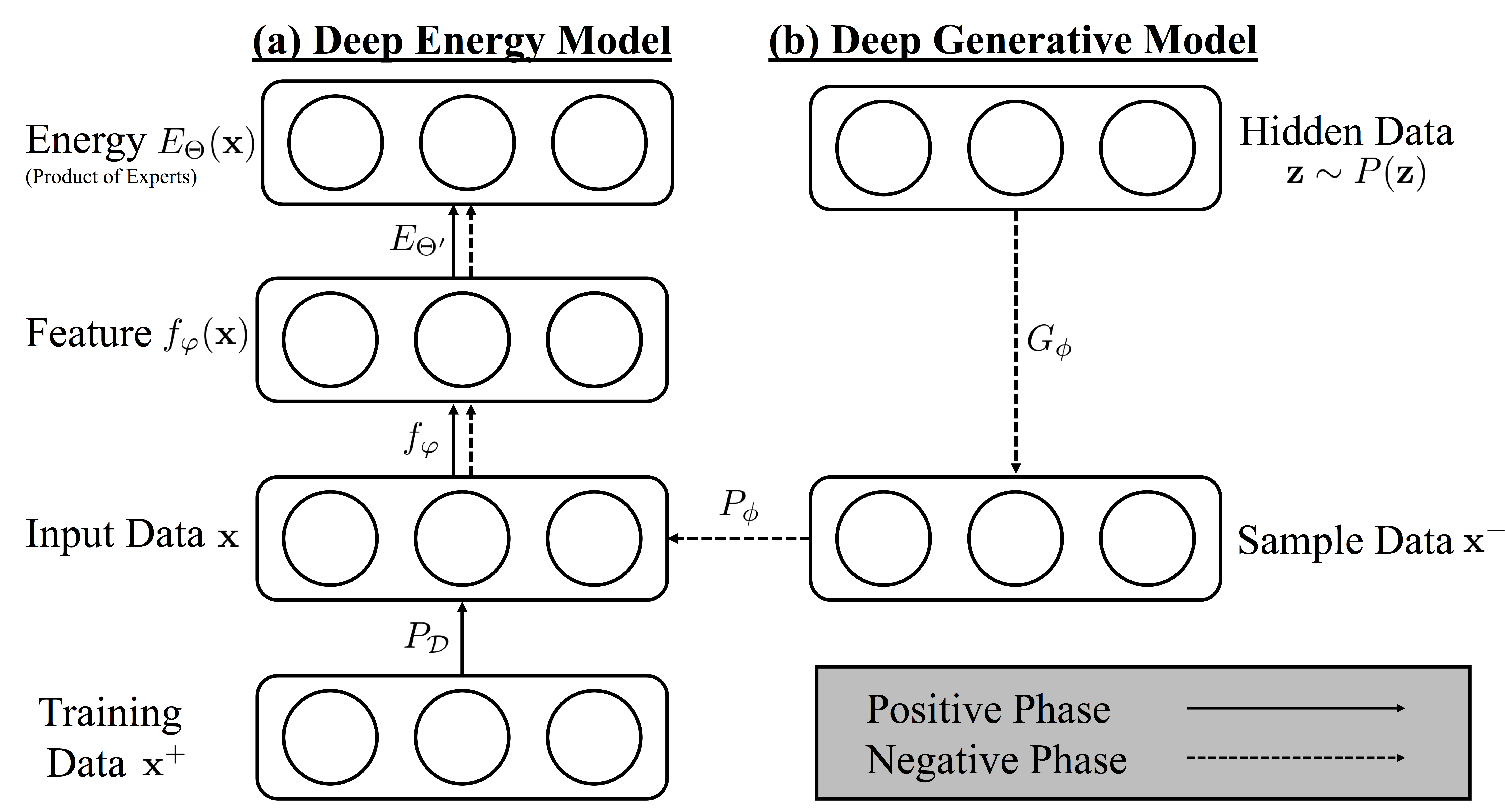}
\caption{The proposed framework has two models that represent two views of what has been learned: (a) a deep energy model is defined to estimate the probability distribution by learning an energy function expressed in terms of a feature space, and (b) a deep generative model deterministically generates samples that approximately match the deep energy model. To train the deep energy model, training examples are used to push down on the energy (positive phase), and samples from the deep generative model are used to push up (negative phase). Moreover, the deep generative model is trained by aligning to the deep energy model.}
\label{fig:proposed_model}
\end{figure}
\section{The Proposed Model}\label{SEC:FRAMEWORK}
We propose a new framework to train energy-based probabilistic models, where the information about the estimated probability distribution is represented in two different ways: an energy function and a generator. 
Ideally, they would perfectly match each other, but in practice as they are trained against each other, one can be viewed as an approximation of the corresponding operation (sampling or computing the energy) associated with the other.
We use only deep neural networks to represent both two models to avoid the need of explicit latent variables and inference over them as well as MCMC sampling. 

The two models are (a) the \textit{deep energy model} (DEM), defining an energy function $E_{\Theta}$ (Eq.~\ref{EQ:ENERGY}), and (b) the \textit{deep generative model} (DGM), a sample generator $G_{\phi}$ trained to match the deep energy model. 
It is important to make sure that two models are approximately aligned during training since they approximately represent two views of what has been learned. 
We present below the training objective for each of these two models. The first update rule is based on Eq.~\ref{EQ:POS_NEG_PHASES}, i.e., approximating the maximum likelihood gradient:
\begin{equation}\label{EQ:TRAIN_DEM}
\text{minimize }\mathrm{D}_{KL}\big(P_{\mathcal{D}}(\mathbf{x})||P_{\Theta}(\mathbf{x})\big) \qquad\text{with negative samples } \mathbf{x}^{-}\sim P_{\phi}(\mathbf{x}) \approx P_{\Theta}(\mathbf{x})
\end{equation}
where $P_{\phi}$ is the sampling distribution of the deep generative model. 
This is mainly to train the deep energy model using negative samples from the deep generative model instead of using an MCMC on $P_{\Theta}$. 
The second update rule is to train the deep generative model to be aligned to the deep energy model, and this also ensures the deep generative model to generate samples appropriately as the data distribution:
\begin{equation}\label{EQ:TRAIN_DGM}
\text{minimize }\mathrm{D}_{KL}\big(P_{\phi}(\mathbf{x})||P_{\Theta}(\mathbf{x})\big) 
\end{equation}
We can view these update rules as resulting in making the three different distributions approximately aligned: $P_{\mathcal{D}}(\mathbf{x}) \approx P_{\Theta}(\mathbf{x}) \approx P_{\phi}(\mathbf{x})$.

\subsection{Deep Energy Model}\label{SEC:DEM}
The energy function $E_{\Theta}$ assigns a scalar energy value to a given input configuration $\mathbf{x}$. 
It represents an energy model distribution $P_{\Theta}$ that  estimates the data distribution $P_{\mathcal{D}}$. 
Interestingly, we can view our deep energy model as a conventional deep classification model that consist of a feature extractor and a discriminator. 
First, the feature extractor $f_{\varphi}$ works on a deep neural-network only to extract high-level features from the input data $\mathbf{x}$. 
Then, the energy function is expressed in terms of the features for capturing higher-order interactions and also in terms of $\mathbf{x}$ and $\mathbf{x}^T\mathbf{x}$ to capture mean and variance over the input data.
We use the form of a product of experts that is analogous to the free energy of an RBM with Gaussian visible units~\cite{DEM}:
\begin{equation}\label{EQ:ENERGY_FUNCTION}
E_{\Theta}(\mathbf{x})=E_{\Theta'}(\mathbf{x}, f_{\varphi}(\mathbf{x}))
=\frac{1}{\sigma^{2}}\mathbf{x}^{T}\mathbf{x}-\mathbf{b}
^{T}\mathbf{x}-\sum_{i}{\log{\big(1+e^{W_{i}^{T}f_{\varphi}(\mathbf{x})+b_{i}}\big)}}
\end{equation}
The first two terms capture the mean and global variance, and the last term is a set of experts $\tilde{E}_{\theta_{i}}(f_{\varphi}(\mathbf{x}))=-\log{\big(1+e^{W_{i}^{T}f_{\varphi}(\mathbf{x})+b_{i}}\big)}$ over the feature data space $f_\varphi(\mathbf{x})$.
Moreover, the integrability of unnormalized distribution $e^{-E_{\Theta}(\mathbf{x})}$ with respect to $\mathbf{x}$ is guaranteed, as each expert grows linearly with using bounded $f_{\varphi}(\mathbf{x})$ and is dominated by the term with $\mathbf{x}^{T}\mathbf{x}$.
Instead of interpreting these experts as latent variables, we use a single-layer neural network to compute an energy value deterministically, and propose a new approach to train it without any iterative sampling methods. 
As we train this model by using the update rule in Eq.~\ref{EQ:POS_NEG_PHASES} and ~\ref{EQ:TRAIN_DEM}, we approximate the negative phase with using samples generated from our deep generative model $G_{\phi}$, as depicted in Fig. \ref{fig:proposed_model}:
\begin{equation}\label{EQ:DEM_APPROX}
\begin{aligned}
\underbrace{\mathrm{E}_{\mathbf{x}^{-}\sim P_{\Theta}(\mathbf{x})}{\Bigg[\dfrac{\partial E_{\Theta}(\mathbf{x}^{-})}{\partial \Theta}\Bigg]}}_{\text{Negative Phase}}&\approx\mathrm{E}_{\mathbf{x}^{-}\sim P_{\phi}(\mathbf{x})}{\Bigg[\dfrac{\partial E_{\Theta}(\mathbf{x}^{-})}{\partial \Theta}\Bigg]}=\mathrm{E}_{\mathbf{z}\sim P(\mathbf{z})}{\Bigg[\dfrac{\partial E_{\Theta}(G_{\phi}(\mathbf{z}))}{\partial \Theta}\Bigg]}\\
&\approx\frac{1}{N}\sum_{i=1}^{N}{\dfrac{\partial E_{\Theta}(G_{\phi}(\mathbf{z}_{i}))}{\partial \Theta}\qquad \text{where } \mathbf{z}_{i}\sim P(\mathbf{z})}
\end{aligned}
\end{equation}
where $\mathbf{z}$ is the latent variable associated with the deep generative model.  
Interestingly, this update rule is exactly the same as explained in Sec.~\ref{SEC:CLASSIFICATION} that the energy function can be considered as a strongly regularized binary classifier to separate samples from two different sources (positive samples from the training dataset $\mathbf{x}^{+}\sim P_{\mathcal{D}}(\mathbf{x})$ and negative samples from the deep generative model $\mathbf{x}^{-}\sim P_{\phi}(\mathbf{x})$).
 
\subsection{Deep Generative Model}\label{SEC:DGM}
The deep generative model has two purposes: provide an efficient non-iterative way of obtaining samples and help training of the energy function by providing approximate samples for its negative phase component of the gradient.  
The sample generating procedure is simple as it is just ancestral sampling from a 2-variable {\em directed graphical model} with a very simple structure: (a) We first sample the i.i.d. latent variable $\mathbf{z}$ from a simple fixed prior $P(\mathbf{z})$, e.g. a uniform sample from $\mathit{U}(-1, 1)$. (b) Then, it is fed into the model $G_{\phi}$, which is a deep neural network, that deterministically outputs samples. 
Moreover, the sampling distribution can be defined by $P_\phi(\mathbf{x})=\int_\mathbf{z} P_\phi(\mathbf{x}|\mathbf{z}) P(\mathbf{z})d\mathbf{z}$, where $P_\phi(\mathbf{x}|\mathbf{z})=\delta_{\mathbf{x}=G_\phi(\mathbf{z})}$ is a Dirac distribution at $\mathbf{x}=G_\phi(\mathbf{z})$, i.e., $\mathbf{x}$ is completely determined by $\mathbf{z}$.

As shown in Eq.~\ref{EQ:TRAIN_DGM}, we train this model to have a similar distribution as the energy model distribution by minimizing the KL divergence between two distributions:
\begin{equation}\label{EQ:DGM_KL}
\mathrm{D}_{KL}\big(P_{\phi}(\mathbf{x})||P_{\Theta}(\mathbf{x})\big)= \mathrm{E}_{\mathbf{x}^{-} \sim P_{\phi}(\mathbf{x})}\big[-\log{P_{\Theta}(\mathbf{x}^{-})}\big]-H\big(P_{\phi}(\mathbf{x})\big)
\end{equation}
The first term is to maximize the log-likelihood of the samples from the generator under the deep energy model. 
It encourages the generator to produce sample with low energy. 
Interestingly, its gradient with respect to $\phi$ does not depend on the partition function of $P_{\Theta}$, and it is computed by using Monte Carlo samples from the generator (as in Eq.~\ref{EQ:DEM_APPROX}) and back-propagating through the energy function: 
\begin{equation}\label{EQ:DGM_APPROX}
\begin{aligned}
\dfrac{\partial}{\partial\phi}\mathrm{E}_{\mathbf{x}^{-} \sim P_{\phi}(\mathbf{x})}\big[-\log{P_{\Theta}(\mathbf{x}^{-})}\big]&=\dfrac{\partial}{\partial\phi}\mathrm{E}_{z\sim P(z)}{\big[-\log{P_{\Theta}(G_{\phi}(z))}\big]}\\
&=\mathrm{E}_{z\sim P(z)}{\Bigg[\dfrac{\partial E_{\Theta}(G_{\phi}(z))}{\partial \phi}\Bigg]}\\
&\approx\frac{1}{N}\sum_{i=1}^{N}{\dfrac{\partial E_{\Theta}(G_{\phi}(z_{i}))}{\partial \phi}\quad \text{ where } z_{i}\sim P(z)}
\end{aligned}
\end{equation}
If we were to only consider the first term of the KL divergence, the generated samples could all converge toward one or more local minima on the energy surface corresponding to the modes of $P_\Theta$. 
The second term in Eq.~\ref{EQ:DGM_KL} is necessary to prevent it that maximizes the entropy of the deep generative model to have just the right amount of variability across the generated samples. 
This is analytically intractable, however, we found that a particular form of regularizer with batch normalization~\cite{BN} can work well to approximately constrain or maximize the entropy.
The batch normalization maps each activation $a_{i}$ into an approximately normal distribution with trainable mean-related shift parameter $\mu_{a_{i}}$ and variance-related scale parameter $\sigma_{a_{i}}$, and the entropy of the normal distribution over each activation can be measured analytically $H(\mathcal{N}(\mu_{a_{i}},\sigma_{a_{i}}))=\frac{1}{2}\log{(2e\pi\sigma_{a_{i}}^2)}$.
We assume this (internal) entropy could effect the entropy of the generator (external) distribution $H(P_{\phi}(\mathbf{x}))$, and therefore approximate it as:
\begin{equation}\label{EQ:ENTROPY}
H(P_{\phi}(\mathbf{x}))\approx \sum_{a_{i}}H(\mathcal{N}(\mu_{a_{i}},\sigma_{a_{i}}))=\sum_{a_{i}}\frac{1}{2}\log{(2e\pi\sigma_{a_{i}}^2)}
\end{equation}
where we sum over all activations in the deep generative model, and this is as regularizing all scale parameters to be increased (contrary to weight decay). 
%Hence, in addition to the energy minimization term, we approximately maximize the entropy of the generated samples by maximizing two times the logarithm of the minibatch scaling factor at the output layer of the generator.

\subsection{Relation to Generative Adversarial Networks}
Generative adversarial networks (GAN)~\cite{GAN} consist of a discriminator $D$ and a generator $G$ trained by a two-player minimax game to optimize the generator so that it generates samples similar to the training examples. 
It motivated our framework based on two separate models, but where the role of the discriminator is played by the energy function.
A big difference and a motivation for the proposed approach is that the GAN discriminator may converge to a constant output as the generator distribution gets closer to the data distribution (the correct answer becomes $D=0.5$ all the time, independently of the input), and potentially forgets a large part of what has been learned earlier along the way (the weights from the input could go to 0, if they were regularized).  
Instead, our energy function can be seen as a discriminator that separates between the training examples and any generator, rather than just again the current generator. Thus, the equivalent discriminator associated with our energy function does not converge to a trivial classifier as the generator improves. 
Of course, another advantage of the proposed approach over GAN is that at the end of training we obtain a generic energy function that can be used to compare any pair of input configurations ($\mathbf{x}_1$ ,$\mathbf{x}_2$) against each other and estimate their relative probability. In contrast, there is no guarantee that a GAN discriminator continues to provide a meaningful output after the end of GAN training (to compare different $\mathbf{x}$
against each other in terms of their $D$ value), except in regions where the generator distribution and the data distribution differ. Instead, in with the proposed framework, as the generated distribution approaches the data distribution, the energy function does not become constant: its gradient becomes constant, meaning that it does not change anymore.

\begin{figure}[htpb]
\centering
\includegraphics[width=1.0\textwidth]{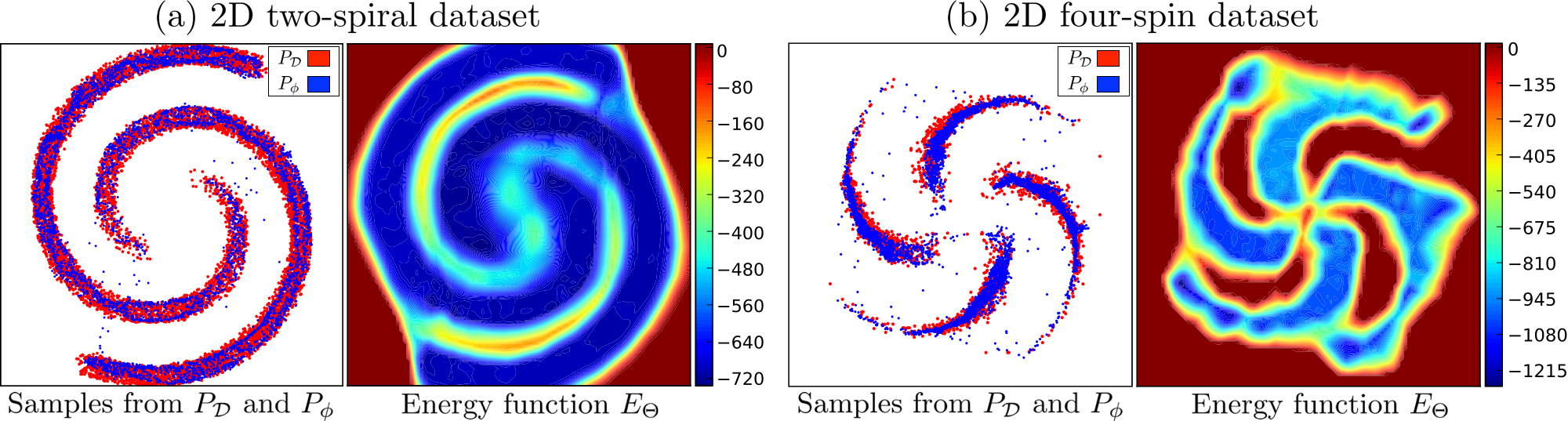}
\caption{Results on 2D datasets to verify visually that all modes are represented, that the energy model matches the data and that the generated samples match the energy function, (a) four-spin dataset and (b) two-spiral dataset. Left : Samples from the training dataset (red) and our deep generative model (blue). Right : Heat map showing the energy function from our deep energy model with blue indicating low energy and red high energy.}
\label{fig:2d_results}
\end{figure}
\begin{figure}[t!]
\centering
\includegraphics[width=1.0\textwidth]{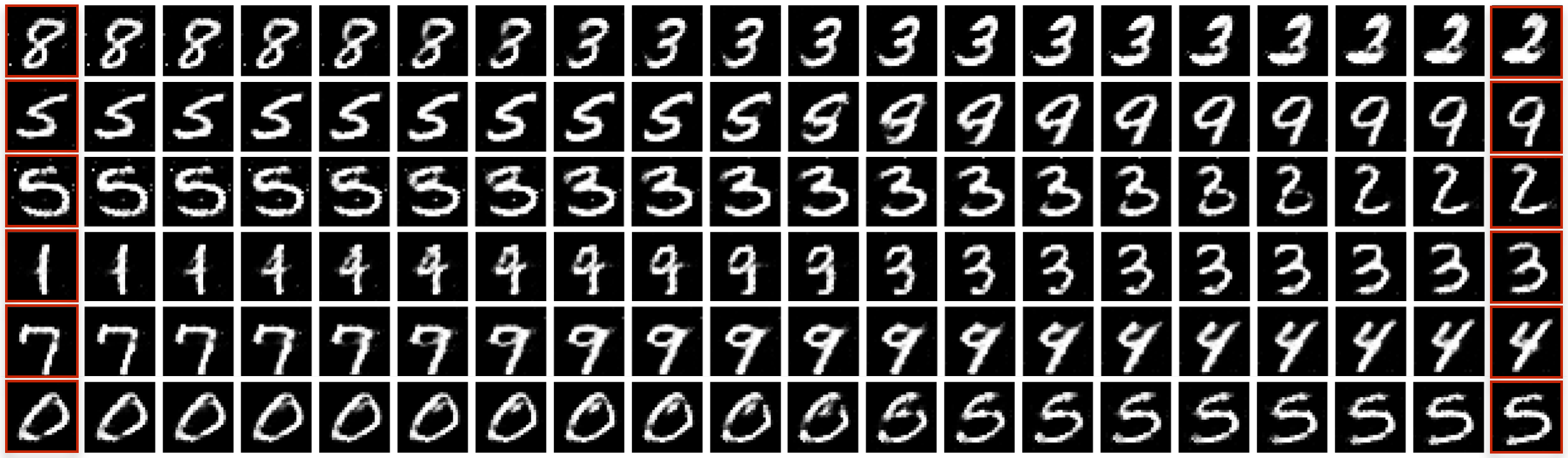}
\caption{Visualization of samples generated from our deep generative model with MNIST, in the leftmost and rightmost columns. The other columns show images generated by interpolating between them in $\mathbf{z}$ space. This model is with only fully-connected layers. Note how the model needs to go through the 3-manifold in order to go from an 8 to a 2, and it goes through a 3 to go from a 5 to a 2, and through a 9 to go from a 7 to a 4, all of which make a lot of sense.}
\label{fig:mnist_results}
\end{figure}
\begin{figure}[htpb]
\centering
\includegraphics[width=1.0\textwidth]{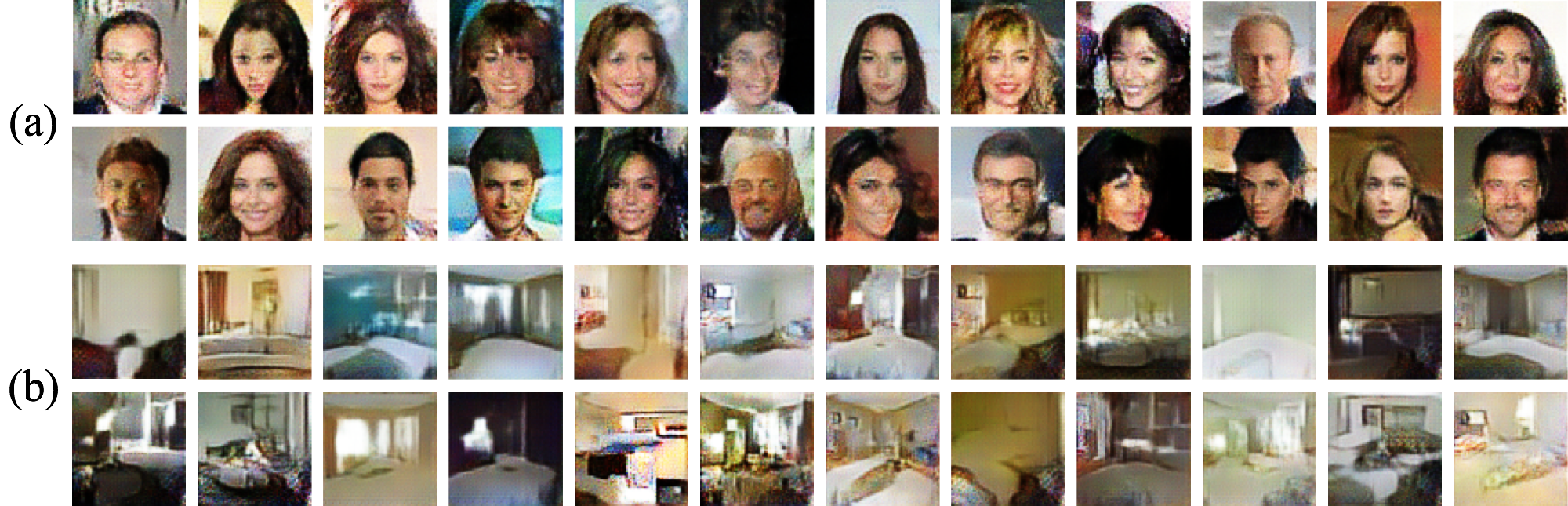}
\caption{Samples generated from the deep generative model with convolutional operations, and trained with 64x64 color-images: (a) CelebA (faces), (b) LSUN (bedroom)}
\label{FIG:FACE_SAMPLES}
\end{figure}

\section{Experiments}
We first experimented our proposed model with 2D-synthetic datasets to show that the deep energy model and the deep generative model are properly learned and matched each other. We generated two types of datasets, the \textit{two-spiral} and \textit{four-spin} datasets. Each dataset has 10,000 points randomly generated under the chosen distribution. For simplicity, we used the same model structure with fully-connected layers for both models, but in reverse orders (ex. the number of hidden units in DEM : 2-128-128-4, DGM : 4-128-128-2, 4 experts, 4 dimensional hidden data $\mathbf{z}$). We used the \textit{AdaGrad}\cite{ADAGRAD} optimizer to train both models with mini-batch stochastic gradient descent. The batch normalization was used only for the deep generative model as the entropy regularizer (Eq.~\ref{EQ:ENTROPY}). Fig.~\ref{fig:2d_results} shows the generated samples as well as the energy function surface.
It can be observed that the deep generative model properly draws samples according to the energy model distribution, and also the energy function fits well the training dataset.
These results show how the proposed framework appropriately aligns two separate models as a single model.

The next evaluation is with the MNIST dataset to learn to unconditionally generate hand-written digits.
It is also trained with using only fully-connected layers (no convolutions or image transformations) for both models. However, we set the number of experts and the hidden data size differently (ex. 128 experts and 10 dimensional hidden data $\mathbf{z}$). 
It is shown in Fig.~\ref{fig:mnist_results} that the deep generative model smoothly generates different samples as the hidden data $\mathbf{z}$ changes linearly.

We also trained the model on high-dimensional data, especially on color images, with using appropriate convolutional operations and structures proposed by \cite{DCGAN}. We used two types of datasets called CelebA~\cite{FACE} and LSUN (bedroom)~\cite{LSUN}. The CelebA is a large-scale face dataset with more than 200K images, and the LSUN is a dataset related to scene categories that especially the bedroom category has nearly 3M images. We cropped and resized both dataset images into 64x64. We used the same model structure with these two datasets that has 1024 experts to define the energy function, and 100 dimensional hidden data $\mathbf{z}$ to generate samples. Both the deep energy model and the deep generative model have 4 layers with convolutional operations (ex. the number of feature maps in DEM: 128-256-512-1024 , DGM: 1024-512-256-128), and all filters are defined by 5x5. The generated samples are visualized in Fig.~\ref{FIG:FACE_SAMPLES}, and this shows how the deep generative model is properly trained. Furthermore, we can also assume that the deep energy model is also trained well as it is the target distribution to estimate.
\section{Conclusion}
The energy-based probabilistic models have been broadly used to define generative processes with estimating the probability distribution. In this paper, we showed that the intractability can be avoided by using two separate deep models only using neural networks. In future work, we are interested in explicitly dealing out with the entropy of generators, and to extend the deep energy model to be used in semi-supervised learning. Moreover, it would be useful to approximately visualize the energy function on high-dimensional input data.

%\subsubsection*{Acknowledgments}
%Use unnumbered third level headings for the acknowledgments. All acknowledgments go at the end of the paper. Do not include acknowledgments in the anonymized submission, only in the final paper.

\bibliography{nips_2016_ref}
\bibliographystyle{plain}

\end{document}